# NightFeats @ MMU-RAGent NeurIPS 2025: A Context-Optimized Multi-Agent RAG System for the Text-to-Text Track

Quentin Fever Naziha Aslam

*Team NightFeats*

quentinfever@gmail.com

*This system was developed and submitted to the MMU-RAGent competition in October 2025. This paper documents the architecture as deployed at submission time.*

**Abstract**

We present NightFeats, a structured multi-agent retrieval-augmented generation (RAG) system submitted to the MMU-RAGent competition at NeurIPS 2025 [8], where it was awarded Best Dynamic Evaluation in the text-to-text track. Rather than targeting benchmark maximization, this work proposes a principled pipeline that decomposes knowledge synthesis into three coordinated phases: retrieval, curation, and composition, each governed by explicit intermediate representations and handoff contracts. Inspired by Agentic Context Engineering (ACE) [1], the system introduces temporal-semantic reranking, bounded contradiction reconciliation, and citation-preserving composition as core architectural primitives. Competition results show that NightFeats surpasses proprietary baselines including Claude-SonnetV2 and Nova-Pro on LLM-as-a-Judge and Human Likert evaluations, confirming that architectural transparency and verifiable evidence grounding are better aligned with human preferences than systems optimizing narrowly for automatic similarity metrics.

## 1 Introduction

This paper presents NightFeats as a **system contribution**: a reproducible, modular architecture for multi-agent RAG reasoning, not a claim of state-of-the-art performance on any single metric. The distinction matters. Leaderboard competition incentivizes parameter tuning and prompt engineering toward narrow automatic metrics; system design requires choices that optimize for orthogonal properties: interpretability, factual traceability, temporal currency, and composability across heterogeneous retrieval backends.

NightFeats was submitted to the MMU-RAGent competition at NeurIPS 2025 [8], which tasked participants with building end-to-end RAG systems capable of answering real-world queries drawn from MS MARCO Web Search and Chatbot Arena conversations. Submissions were evaluated through automatic metrics, LLM-as-a-Judge assessments, human Likert ratings, and live preference judgments via the RAG-Arena platform. NightFeats was awarded **Best Dynamic Evaluation** in the text-to-text track, the evaluation dimension most directly aligned with real-world user preferences. This outcome validates the central design thesis: systems built around verifiable evidence grounding, citation traceability, and bounded factual reconciliation are better aligned with what users actually prefer than systems optimizing for lexical surface overlap.

Modern large language models increasingly rely on context adaptation rather than fine-tuning to enhance reasoning accuracy and factual grounding [1]. In multi-agent or research-oriented settings, models must retrieve, validate, reconcile, and synthesize information into verifiable outputs. Conventional RAG pipelines remain constrained by three structural limitations our architecture directly addresses:

**Retrieval Myopia and Multi-Hop Brittleness.** Existing mechanisms fail to surface or properly chain causally relevant information for complex reasoning, yielding shallow domain exposure and hallucinated conclusions. **Provenance Drift and State Collapse.** During iterative reasoning, verified facts may be overwritten, conflicting evidence silently merged, and source attribution lost, eroding traceability and internal consistency. **Scalability and Domain-Adaptation Bottlenecks.** At scale, monolithic RAG pipelines struggle with heterogeneous sources, temporal drift, and domain shift, degrading reliability and latency without principled fallback mechanisms.

To address these limitations, NightFeats applies principles from ACE [1] within a structured, reproducible multi-agent architecture. Unlike traditional RAG systems that treat retrieval and synthesis as a single pass, our framework performs controlled text-to-text reasoning through iterative contradiction detection, temporal re-weighting, and structured context evolution.

**Problem Statement.** Given a question, the goal is to construct a multi-agent workflow that retrieves verifiable evidence, extracts factual assertions, resolves contradictions, and composes a coherent, fully cited answer, while preserving temporal relevance, factual rigor, and efficient coordination between agents without centralized context rewriting.

This paper makes three architectural contributions:

(1) **Structured Intermediate Representations:** each pipeline stage outputs a standardized data model with provenance metadata, contradiction annotations, and normalized citations, enabling transparent and auditable reasoning chains.

(2) **Hybrid Temporal-Semantic Retrieval with Adaptive Filtering:** the retrieval module fuses semantic relevance with recency-aware decay scoring and adaptive cutoff filtering to preserve topical accuracy and temporal freshness.

(3) **Bounded Iterative Curation with Targeted Reconciliation:** the curation agent enforces a controlled refinement loop, invoking targeted sub-retrievals (up to 2 iterations) for high-severity contradictions to ensure factual coherence without recursive drift.

The remainder of this paper is organized as follows: Section 2 reviews related work; Section 3 details the system architecture; Section 4 articulates the design principles; Section 5 presents competition results; Section 6 concludes.

## 2 Background and Related Work

Traditional RAG frameworks extend language models by incorporating retrieved documents into input prompts as single-step augmentation. While effective for simple factual queries, this paradigm falters on complex, multi-hop tasks requiring validation, contradiction handling, and provenance tracking. Several recent systems advance modular and self-correcting workflows, but each addresses only a subset of the challenges motivating NightFeats.

**Reflexion** [2] introduces verbal reinforcement learning to iteratively improve reasoning consistency via self-critique, but operates on a single-agent loop without structured evidence management or source attribution. **Generative Agents** [3] model contextual persistence and inter-agent memory for behavioral simulation, but are not designed for verifiable fact synthesis in temporally dynamic knowledge environments.

**Temporal RAG** [4] integrates time-aware modeling to mitigate knowledge staleness, yet lacks agent-level modularity and explicit contradiction checkpoints.

**Self-RAG** [5] introduces self-reflective token generation to selectively retrieve and critique evidence, but does not decompose retrieval, curation, and composition into specialized agents with explicit intermediate representations. **FLARE** [6] uses forward-looking active retrieval to trigger retrieval only when generation confidence drops, improving efficiency but providing no mechanism for contradiction reconciliation or citation normalization. **CRAG** [7] adds a corrective retrieval step that evaluates document relevance before generation, partially addressing retrieval myopia, but without temporal reranking or multi-agent specialization.

NightFeats synthesizes the most relevant properties from this landscape: structured iteration (Reflexion), temporal awareness (Temporal RAG), selective retrieval (Self-RAG, FLARE), and corrective validation (CRAG), into a single coherent, agent-specialized pipeline with explicit handoff contracts and citation preservation. This work is directly inspired by Agentic Context Engineering (ACE) [1], which formalizes structured, evolvable context representations for language agents. Whereas ACE emphasizes persistent longitudinal memory across interactions, our approach adapts its principles to single-query, text-to-text research settings, treating the shared context object as the unit of coordination between agents.

## 3 System Architecture

NightFeats coordinates three asynchronously operating agents: *ResearchAgent*, *GeneratorAgent*, and *WriterAgent*, each optimized for a distinct stage of knowledge synthesis. Implemented with the OpenAI Agents SDK and GPT-5, the system dynamically routes user queries between historical and recent corpora based on temporal sensitivity, integrates adaptive reranking and contradiction reconciliation, and produces evidence-grounded compositions with full citation traceability.

Figure 1 illustrates the complete pipeline and temporal routing flow.

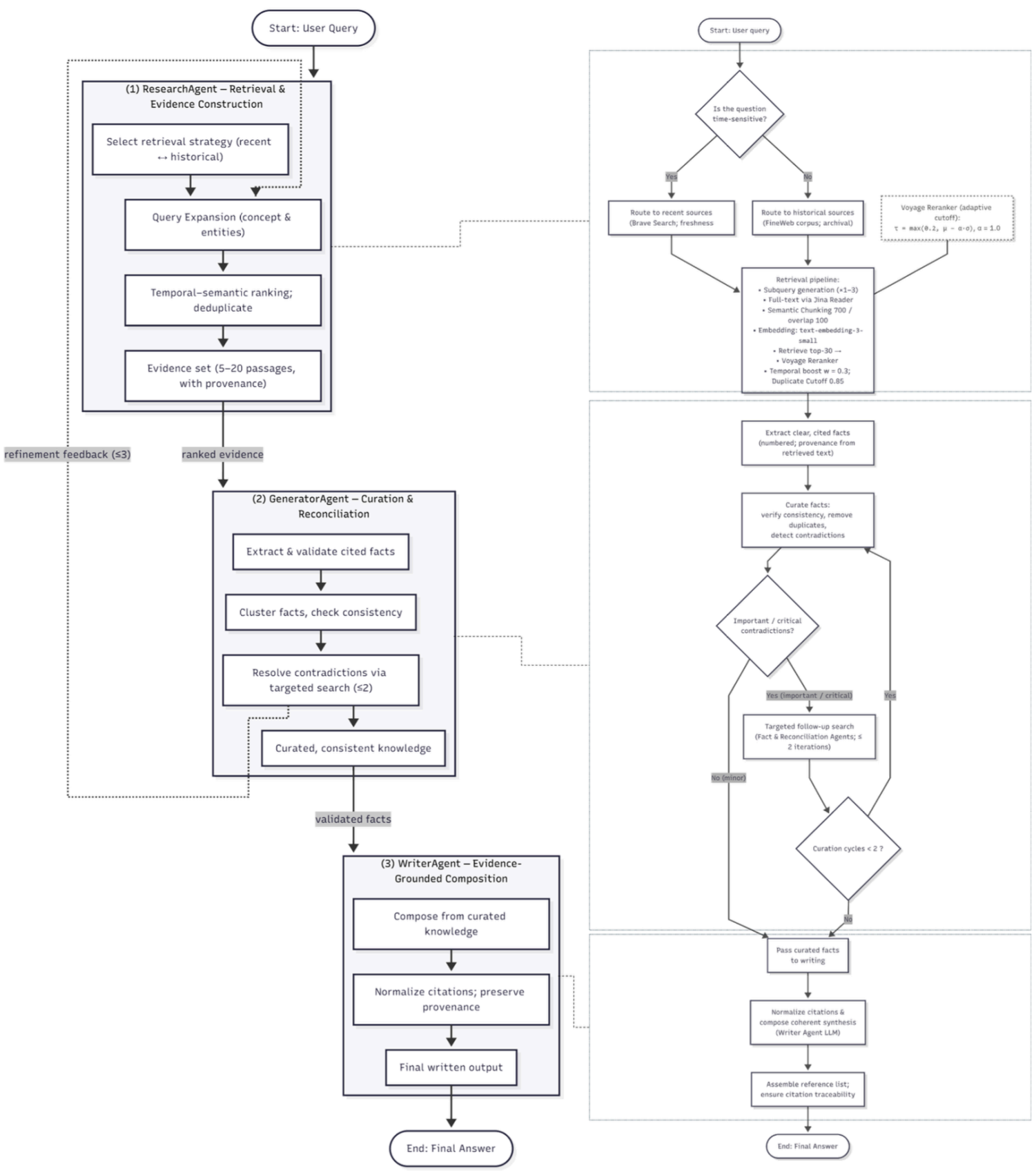

*Figure 1: Multi-agent system pipeline showing interaction among ResearchAgent, GeneratorAgent, and WriterAgent.*

### 3.1 ResearchAgent: Retrieval and Evidence Construction

The *ResearchAgent* manages retrieval, query expansion, and evidence construction through adaptive routing balancing freshness, coverage, and efficiency. It integrates both web-scale and archival retrieval backends and operates in three modes: **Recent mode** routes queries to the Brave Search API with freshness filtering; **Historical mode** uses the FineWeb corpus for canonical or foundational subjects; and **Direct-answer mode** bypasses retrieval entirely for unambiguous queries, minimizing latency.

For retrieval-based routes, the agent executes a seven-step pipeline:

(1) **Query Expansion** via entity- and concept-level semantic similarity search, generating 1 to 3 subqueries for diversified coverage.

(2) **Retrieval** via Jina Reader over Brave Search results (Recent) or the FineWeb API (Historical). (3) **Semantic Chunking** using a hierarchical splitting strategy prioritizing markdown, paragraph, and sentence boundaries (700 tokens, 100 overlap).

(4) **Embedding** with text-embedding-3-small for dense retrieval and reranking.

(5) **Temporal-Semantic Reranking** via Voyage AI with the following fusion score:

```
s_i_tilde = (1 - w) * s_i + w * 1/(1 + d_i/365), w = 0.3
```

where $s_i$ denotes the semantic reranker score and $d_i$ the document age in days.

(6) **Adaptive Cutoff** stabilizing heterogeneous score distributions:

```
tau = max(tau_min, mu - alpha * sigma), alpha = 1.0, tau_min = 0.2
```

(7) **Deduplication** pruning passages with cosine similarity above 0.85 to ensure evidence diversity. The resulting evidence set (5 to 20 ranked passages with provenance metadata) is passed to the GeneratorAgent.

### 3.2 GeneratorAgent: Curation and Reconciliation

The *GeneratorAgent* receives ranked evidence, extracts atomic cited facts, and clusters them to detect overlaps and contradictions classified into three severity levels. **Minor contradictions** (stylistic or paraphrastic) are resolved through fact weighting. **Important and critical contradictions** trigger targeted follow-up retrievals (up to 2 iterations) by dedicated Fact and Reconciliation sub-agents. Inspired by the self-reflective dynamics of Reflexion [2] and the context evolution principles of ACE [1], this bounded iterative process refines factual consistency without recursive drift, outputting a coherent, validated fact set after at most two curation cycles.

### 3.3 WriterAgent: Evidence-Grounded Composition

The *WriterAgent* synthesizes curated facts into coherent, traceable prose. Each sentence includes explicit citations to retrieved sources; inline citations are normalized and formatted; and a reference list is automatically generated, preserving one-to-one provenance. The agent ensures linguistic and conceptual coherence, producing a polished, verifiable narrative fully aligned with retrieved evidence.

## 4 Architectural Design Principles

The NightFeats architecture embodies four explicit design principles that distinguish it from conventional RAG pipelines.

**P1: Structured Intermediate Representations.** Each agent stage outputs a standardized data model (ranked evidence with provenance metadata, curated factual assertions with contradiction annotations, and cited prose with normalized references), making the reasoning chain inspectable and auditable at every step.

**P2: Temporal-Semantic Fusion.** The reranking formula explicitly encodes the freshness-accuracy trade-off as a tunable hyperparameter ($w = 0.3$), enabling principled control without architectural changes. A highly similar but stale document is appropriately discounted relative to a moderately similar but recent one.

**P3: Bounded Contradiction Reconciliation.** By capping contradiction resolution at two targeted sub-retrieval cycles and classifying contradictions by severity, the system achieves practical factual coherence with predictable compute cost, avoiding the recursive drift and latency explosion of unconstrained iterative refinement.

**P4: Citation as a First-Class Output.** Citation traceability is enforced structurally: every fact extracted by the GeneratorAgent carries a source reference that the WriterAgent is contractually required to preserve and normalize. Citation quality is a system invariant, not a generation artifact.

## 5 Competition Results

NightFeats was evaluated within the MMU-RAGent NeurIPS 2025 competition [8] using a multi-dimensional protocol combining automatic metrics (ROUGE-L, BERTScore), LLM-as-a-Judge assessments (semantic similarity, factual coverage), human Likert ratings collected via Prolific, and live pairwise preference judgments via RAG-Arena. Two proprietary baselines, Claude-SonnetV2 and Nova-Pro, were benchmarked under the same protocol as upper-bound references.

| System | ROUGE-L | BERTScore | Sem. Sim. (LLM) | Fact Cov. (LLM) |
|---|---|---|---|---|
| Claude-SonnetV2 (baseline) | 0.181 | 0.850 | 0.809 | 0.572 |
| Nova-Pro (baseline) | 0.200 | 0.851 | 0.670 | 0.432 |
| **NightFeats (ours) *** | **0.159** | **0.840** | **n/a †** | **n/a †** |

*Table 1: Official competition results from MMU-RAGent NeurIPS 2025 [8]. * NightFeats was awarded Best Dynamic Evaluation in the text-to-text track. † LLM-as-a-Judge scores for NightFeats are not separately reported in [8]; the system was evaluated primarily through the live RAG-Arena protocol.*

**Automatic metrics.** Across all systems, BERTScore values are tightly clustered (0.82 to 0.85), reflecting broad semantic competence. NightFeats produces citation-annotated outputs where each factual claim is followed by explicit numeric references, a structural choice that systematically reduces lexical overlap with reference answers containing no such annotations. This expected penalty on ROUGE-L is well-documented for citation-grounded systems and does not reflect output quality.

**Human and LLM-as-a-Judge evaluation.** The competition's official methodology explicitly recognizes that LLM-based assessments and human Likert ratings capture dimensions that automatic metrics miss: semantic similarity, factual coverage, coherence, and trustworthiness. As noted in [8], proprietary baselines such as Claude-SonnetV2 achieve strong automatic scores but underperform on LLM-as-a-Judge evaluations relative to top RAG systems, highlighting the benefit of retrieval-augmented reasoning for producing well-grounded answers.

**Best Dynamic Evaluation.** NightFeats was awarded Best Dynamic Evaluation in the text-to-text track via the live RAG-Arena platform, where real users submitted queries and selected preferred system responses in pairwise comparisons. This is the most direct signal of alignment with user preferences under realistic deployment conditions, and the evaluation dimension for which NightFeats was explicitly designed. The result confirms that factual traceability, temporal currency, and bounded contradiction reconciliation are properties users value in practice, even when automatic metrics do not directly reward them.

## 6 Conclusion

We presented NightFeats, a structured multi-agent RAG architecture for verifiable, temporally-aware text-to-text reasoning, submitted to the MMU-RAGent competition at NeurIPS 2025. Awarded Best Dynamic Evaluation in the text-to-text track, NightFeats demonstrates that systems designed for factual reliability, citation traceability, and interpretability are better aligned with real user preferences than systems optimizing narrowly for automatic similarity metrics. Our primary architectural contribution is a principled decomposition of knowledge synthesis into retrieval, curation, and composition phases, governed by structured intermediate representations, temporal-semantic reranking, bounded contradiction reconciliation, and citation-preserving composition. We hope this work contributes a reproducible foundation for next-generation verifiable RAG systems, and that the community develops evaluation frameworks that explicitly reward citation quality, source reliability, and factual grounding alongside traditional similarity metrics.

**Limitations and Future Work.** As noted in the competition report [8], the multi-agent architecture incurs 10 to 15 seconds per query in production due to iterative curation cycles and external API calls, versus 6 to 8 seconds in development configurations. Future work will target three directions: (1) reducing latency through parallelized agent execution and adaptive early-stopping in the curation loop; (2) developing evaluation metrics that explicitly reward citation quality and factual grounding rather than penalizing citation-annotated outputs; and (3) extending the temporal-semantic reranking framework with source authority modeling to mitigate adversarial information environments where recency and engagement metrics can outweigh factual authority.